%% file: main.tex
\documentclass[12pt]{article}

\usepackage{preamp}

\usepackage{times}

\usepackage{graphicx,subcaption}
\usepackage{amsmath}
\usepackage{algorithm}
\usepackage{algpseudocode}
\algnewcommand\algorithmicforeach{\textbf{for each}}
\algdef{S}[FOR]{ForEach}[1]{\algorithmicforeach\ #1\ \algorithmicdo}
\usepackage{blindtext}
\usepackage[hyphens]{url}
\usepackage{hyperref}
\usepackage{ifoddpage}
\title{Versatile modular neural locomotion control with fast learning} 

\author
{Mathias Thor,$^{1\ast}$ Poramate Manoonpong,$^{1,2}$\\
\\
\normalsize{$^{1}$Embodied AI and Neurorobotics Laboratory, SDU Biorobotics,}\\
\normalsize{The M{\ae}rsk Mc-Kinney M{\o}ller Institute, The University of Southern Denmark}\\
\normalsize{Campusvej 55, Odense 5230, Denmark}\\
\\
\normalsize{$^{2}$Bio-Inspired Robotics and Neural Engineering Laboratory,}\\
\normalsize{School of Information Science and Technology, Vidyasirimedhi Institute of Science and Technology}\\
\normalsize{Rayong 21210, Thailand}\\
\\
\normalsize{$^\ast$Correspondence to:  mathias@mmmi.sdu.dk.}
}

\date{}

\begin{document} 

\baselineskip24pt

\maketitle 

\newpage
\paragraph{Abstract:}
Legged robots have significant potential to operate in highly unstructured environments. The design of locomotion control is, however, still challenging. Currently, controllers must be either manually designed for specific robots and tasks, or automatically designed via machine learning methods that require long training times and yield large opaque controllers. Drawing inspiration from animal locomotion, we propose a simple yet versatile modular neural control structure with fast learning. The key advantages of our approach are that behavior-specific control modules can be added incrementally to obtain increasingly complex emergent locomotion behaviors, and that neural connections interfacing with existing modules can be quickly and automatically learned. In a series of experiments, we show how eight modules can be quickly learned and added to a base control module to obtain emergent adaptive behaviors allowing a hexapod robot to navigate in complex environments. We also show that modules can be added and removed during operation without affecting the functionality of the remaining controller. Finally, the control approach was successfully demonstrated on a physical hexapod robot. Taken together, our study reveals a significant step towards fast automatic design of versatile neural locomotion control for complex robotic systems.

\newpage
\input{tex/introduction}
\input{tex/results}
\input{tex/discussion}
\input{tex/methods}

\section*{Data availability} 
All data from running the experiments as well as the learned weight sets can be accessed at \url{https://github.com/MathiasThor/CPG-RBFN-framework/tree/main/data}

\section*{Code availability} 
The source code for running the controller in simulation can be accessed at \url{https://github.com/MathiasThor/CPG-RBFN-framework}

\bibliography{bib}
\bibliographystyle{naturemag}

\section*{Acknowledgments}
The authors would like to thank the members of the SDU Biorobotics group for their technical support and helpful discussions. The authors would also like to thank Anders Lyhne Christensen for fruitful discussions and detailed feedback 
\section*{Funding:} This work was supported in part by the Horizon 2020 Framework Programme (FETPROACT-01-2016 FET Proactive: Emerging Themes and Communities) under Grant 732266 (Plan4Act) (P.M., Project WP-PI) and in part by a startup grant on Bio-inspired Robotics from the Vidyasirimedhi Institute of Science and Technology (VISTEC) (P.M., Project PI). 

\section*{Author contributions} M.T. contributed to the modular neural control structure, implementation, experiments, data analysis, and the manuscript. P.M. contributed to the embodied neural control structure, data analysis, and the manuscript. 

\section*{Competing Interests} The authors declare that they have no competing interests. 

\newpage
\subsection*{Supplementary materials}

\renewcommand{\thealgorithm}{S\arabic{algorithm}}
\setcounter{algorithm}{0}
\renewcommand{\thefigure}{S\arabic{figure}}
\setcounter{figure}{0}

\textbf{This PDF includes:}
\begin{itemize}
  \setlength\itemsep{0.000001em}
  \item Section S1. Simulation implementation
  \item Section S2. Controller generalization and limitations
  \item Section S3. Disabling control modules
  \item Figure S1. Body posture limitations
  \item Figure S2. Learning process.
  \item Figure S3. The physical version of MORF.
  \item Algorithm S1. $PI^{BB}$ Pseudocode
\end{itemize}

\noindent\textbf{Other Supplementary Material for this manuscript includes:}
\begin{itemize}
  \setlength\itemsep{0.000001em}
  \item Video S1 Learning the base controller {\footnotesize (\url{https://youtu.be/wBMH6HuDTms})}
  \item Video S2 Learning the obstacle reflex controller {\footnotesize (\url{https://youtu.be/VFxM8FNHwlk})}
  \item Video S3 Learning the body posture controller {\footnotesize (\url{https://youtu.be/LpWXVrbPj38})} 
  \item Video S4 Learning the directional locomotion controller {\footnotesize (\url{https://youtu.be/-8k73wVa89E})} 
  \item Video S5 Deploying primitive controllers in simulation {\footnotesize (\url{https://youtu.be/MRSWHOnPZnQ})} 
  \item Video S6 Deploying primitive controllers on a physical robot {\footnotesize (\url{https://youtu.be/w1T2uxM_4KE})} 
  \item Video S7 Deploying primitive and advanced controllers in simulation {\footnotesize (\url{https://youtu.be/uzQN6vsHuww})}
  \item Video S8 Controller generalization and limitations {\footnotesize (\url{https://youtu.be/7aX7aqxLOs0})} 
  \item Video S9 Disabling primitive control modules {\footnotesize (\url{https://youtu.be/NkntPiiMoRU})} 
  \item Video S10 Disabling advanced control modules {\footnotesize (\url{https://youtu.be/X9e08AHq7kM})}

\end{itemize}

\subsection*{S1. Simulation details}
For the simulated experiments we use the robot simulation framework CoppeliaSim from Coppelia Robotics with the Vortex physics engine from CM Labs. CoppeliaSim offers real-world parameters (i.e., corresponding to physical units) for many physical properties, making it both realistic and precise. However, to make the learned controller more robust, Gaussian noise is added to the center of mass for MORF. Since MORF is mostly rigid, we saw no need to introduce additional simulation noise.

\subsection*{S2. Controller generalization and limitations}
In the following, we investigate the generalization and limitations of the learned control modules. All of the generalizations and limitations are furthermore shown visually in Supplementary Video S8.

\subsubsection*{Obstacle reflex control}
Using the obstacle reflex controller, MORF is able to climb up obstacles. The obstacle reflex controller presented in this work is learned on obstacles with a height of $0.04\,$m, but it can generalize to obstacles with a max height of $0.07\,$m (i.e., a 75\% increase). The obstacle reflex controller can be combined with the directional locomotion controller to address the issue of keeping a straight line when the obstacle height is big and the behavior may affect the walking direction of the robot. As we will later explain, the body posture controller furthermore generalizes to the high and low locomotion behaviors. 

\subsubsection*{Body posture control}
\begin{figure}[h]
\centering
\includegraphics[width=0.6\textwidth]{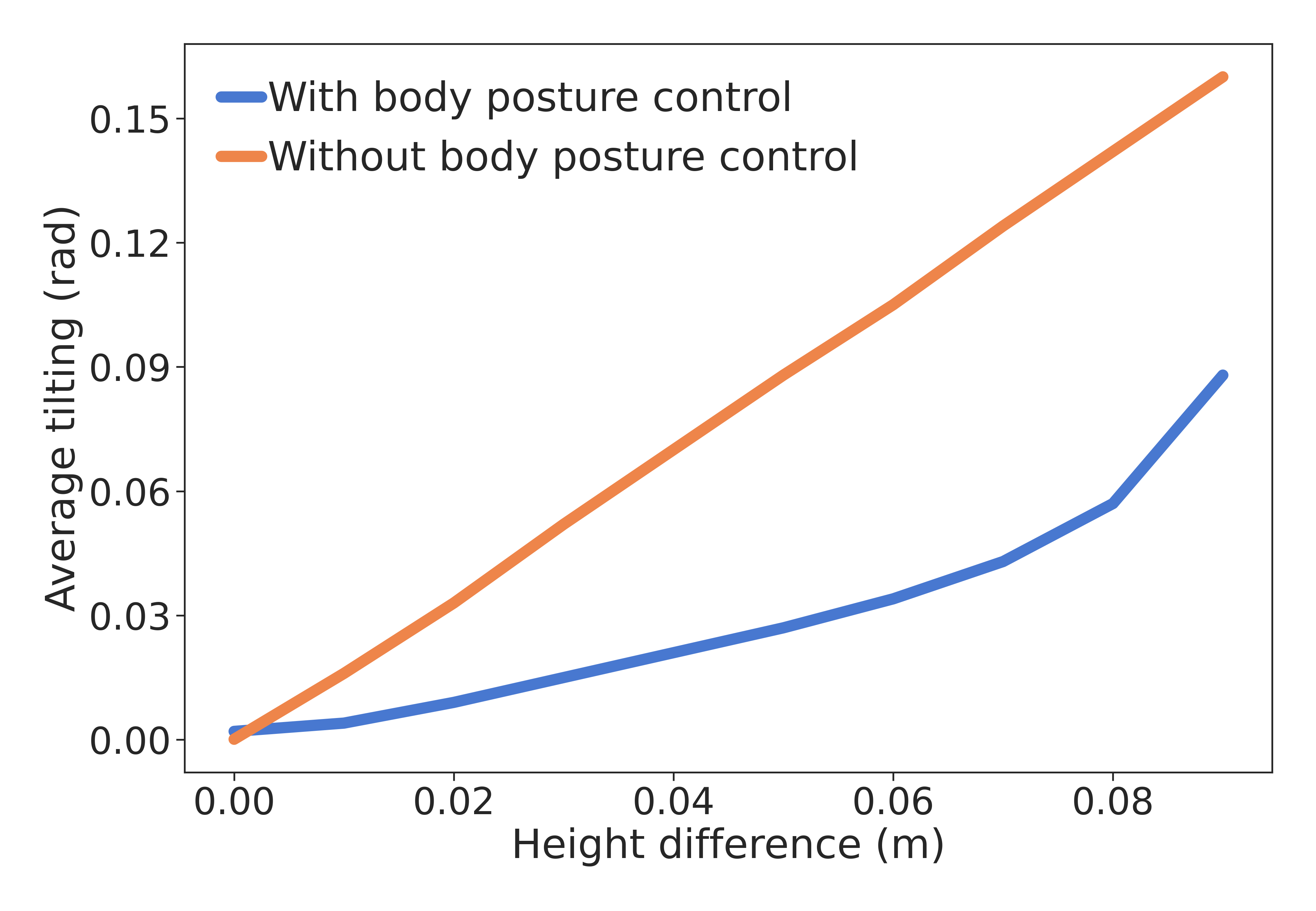}
  \caption{\small \textbf{Body posture limitations}. The average tilting of MORF when walking for 5 seconds with the right side legs on a plate of different heights (i.e., height differences). The plot shows both the average tilt with and without the body posture controller.}
  \label{fig:tilt_limit}
\end{figure}

Using the body posture controller, MORF is able to minimize tilt movement while walking. The body posture controller presented in this work can generalize to various differences in walking heights although learned with a fixed height difference of $0.04\,$m. Fig.\ \ref{fig:tilt_limit} shows how the relationship between the difference in walking height and the amount of tilting when MORF is walking for 5 seconds with its right legs on the levitated plate. From this, it can be seen that the body posture controller can keep the body tilting low when compared to not using the controller. At around a height difference of $0.08\,$m, the tilting begins to increase rapidly, and MORF struggles to move straight forwards. A difference in walking height of $0.08\,$m may therefore be considered the controller's limitation. The body posture controller can be combined with the directional locomotion controller to address the issue of keeping a straight line when the difference in walking height is big. Moreover, as we will later explain, the body posture controller also generalizes to high locomotion as well as wall and pipe climbing behaviors were it plays an important role when varying the friction. 

\subsubsection*{Directional locomotion control}
Using the directional locomotion controller, MORF is able to locomote in various directions. The directional locomotion controller presented in this work can generalize to all walking directions, although learned with a fixed desired walking direction of $45$ degrees to the right. As we will later explain, the directional locomotion controller furthermore generalizes to high, low, and narrow locomotion. 

\subsubsection*{High locomotion control}
Using the high locomotion controller, MORF is able to lift its body to clear obstacles in its way. The maximum height and width of the obstacles which can be cleared by MORF when using high locomotion control is $0.08\,$m and $0.29\,$m, respectively. This is an increase of 128.6\% in height when compared to the base control. The maximum dimensions of the obstacle that can be cleared with high locomotion control correspond to the minimum distance from the bottom of the robot to the walking ground and the minimum distance between bilateral legs. 

The high locomotion controller can be combined with the body posture controller for reducing tilt movement. E.g., when MORF is walking for five seconds with a walking height difference of $0.04\,$m, the tilting is reduced by 64\% on average. The high locomotion controller can also be combined with the directional locomotion controller for steering. Experiments showed that it, as with the base controller, can steer in any direction but at a reduced turning speed. Finally, the height walk controller can also be combined with the obstacle reflex controller. In fact, the increased lifting height of the legs during the swing phase when using the high locomotion controller enables MORF to traverse obstacles up to $0.07\,$m in height before requiring the obstacle reflex controller. This creates redundancy that can be exploited in case of sensory failure. With the obstacle reflex controller, the high locomotion controller enables MORF to traverse obstacles up to $0.095\,$m in height. This corresponds to 70.4\% of MORF's body height, and it is 35.7\% taller than the base controller with obstacle reflex control.

\subsubsection*{Low locomotion control}
Using the low locomotion controller, MORF is able to squeeze below obstacles in its way. The minimum height below the obstacle which MORF can fit under using this controller is $0.146\,$m, which is only 8.9\% taller than MORF's body height.  

The low locomotion controller makes MORF lay on its stomach with no legs in ground contact in parts of the step cycle to minimize the height. This prevents the controller from being used in parallel with the body posture controller since when no leg is in ground contact, the body cannot be stabilized. Like the high locomotion controller, the low locomotion controller can also be combined with the directional locomotion controller for steering during low locomotion. It could likewise steer in any direction with a reduced turning speed. The low locomotion controller can, however, not be combined with the obstacle reflex controller. This is because the reduced height of MORF results in the optic distance sensory placed on MORF's head detecting the walking ground. The obstacle reflex is hereby constantly triggered even when no obstacle is in front of MORF. A solution is to reduce the cutoff distance from $0.115\,$m to $0.07\,$m when using the low locomotion controller. When doing so, MORF is able to traverse obstacles up to $0.03\,$m in height.

\subsubsection*{Narrow locomotion control}
Using the narrow locomotion controller, MORF is able to squeeze between obstacles. The minimum width between obstacle is $0.4\,$m which correspond to the maximum width of MORF when the narrow locomotion controller. Compared to using only the base controller, this is 17.5\% narrower. 

The narrow locomotion controller can be combined with the directional locomotion controller for steering. It can steer in any direction with a similar turning speed to when using the base controller. However, the narrow configuration of the legs prevents the controller from being used in parallel with the body posture and obstacles reflex controller.

\subsubsection*{Pipe climb control}
Using the pipe climb controller, MORF is able to locomote on a pipe. The controller was learned on a pipe with a diameter of $0.7\,$m, but the controller generalizes to any pipe diameters above $0.6\,$m. Furthermore, it can deal with a 5\% reduction in pipe friction on a $0.7\,$m diameter pipe (with an initial friction coefficient of 1.0). 

When the pipe climb controller is combined with the body posture controller, the pipe friction can be decreased by up to 99\% (i.e., a friction coefficient of 0.01). The pipe climb controller can not make use of either the directional locomotion or obstacle reflex controller. Although their combinations show strong indications of being able to steer on pipes that bend and climb from one pipe diameter to another, the instability involved in these actions causes the robot to fall off the pipe. A solution would be to use more complex foot designs that incorporate electromagnets or suctions cups, thereby ensuring the robots stay on the pipe.

\subsubsection*{Wall climb control}
Using the wall climb controller, MORF is able to climb forwards between to walls without being in ground contact. The controller was learned on walls that are placed $0.41\,$m apart. With this width, MORF can clear a $1.2\,$m gap on average by climbing on the walls. The controller can generalize to walls that are placed up to $0.435\,$m apart. When the wall are further apart MORF is able to climb further upwards. In future work is would be interesting to exploit this and lean a decent and ascent controller to add on top of the wall climb controller. When the friction is reduced by 20\% (from an initial friction coefficient of 1.0) on both walls, the clearing distance is reduced to $0.82\,$m average. Reducing only the friction on one wall results in a clearing distance of $0.5\,$m.

When the wall climb controller is combined with the body posture controller, the clearing distance is the same when using the initial friction on the walls when compared to not using the body posture controller. However, it is 77.9\% longer when reducing the friction on both walls by 20\% and 42\% longer when only reducing the friction on one wall. When the friction is reduced by 20\% on one wall, the activity of the body posture controller increases by 20.6\% compared to using the initial friction on both walls. Like in the case of the pipe climb controller, neither the directional locomotion nor obstacle reflex controller could be used together with this wall climb controller.

\subsubsection*{General observations}
The above investigation shows that the primitive closed-loop modules generalize to other behaviors besides the one produced by the base controller. In many cases, this results in emergent behaviors with better performance than using a single control module with the base controller. 

In some cases, two control modules could not work in parallel. This was either because the leg configuration of the advanced behavior is too different from that of the base behavior (e.g., the narrow locomotion behavior) or because the advanced behavior is sensitive regarding stability (e.g., the pipe climbing behavior). There are two possible solutions to this. The first is to learn dedicated primitive closed-loop modules for the specific advanced behaviors. The second is to allow the modules to inhibit each other in a hierarchical manner. More important behaviors can, in this way, turn off less important behaviors that interfere negatively (e.g., narrow locomotion inhibits the directional locomotion control module). In the results presented in this work, we employ the inhibition technique where an advanced controller will inhibit non-compliant primitive controllers.

\subsection*{S3. Disabling control modules}
To investigate how the robot performs without specific modules, we disabled each of the control modules in the combined controller experiments. Supplementary Video S9 shows the results from disabling the modules of the controller that uses the three primitive closed-loop controllers (obstacle reflex, body posture, and directional locomotion controller). Movie S9 and S10 shows the results from disabling the modules of the controller that uses the three primitive closed-loop controllers as well as the one that uses both the primitive and advanced controllers. Both videos show that all the learned control modules are needed in order to locomote from the initial to the end position.

\newpage
\begin{figure}[h]
\centering
\includegraphics[width=1\textwidth]{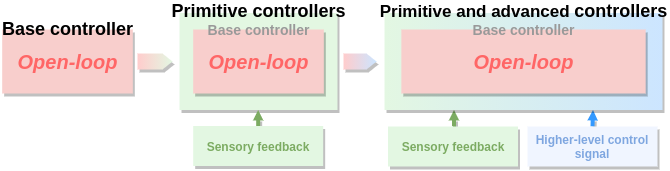}
  \caption{\small \textbf{Learning process}. First the base controller is learned without any feedback. Then the primitive controllers are learned using sensory feedback and added in parallel with the base controller. Finally, the advanced controller are learned using high-level control signals and added in parallel with the base and primitive controllers.}
  \label{fig:learning_approach}
\end{figure}

\begin{figure}[]
\centering
\includegraphics[width=0.75\textwidth]{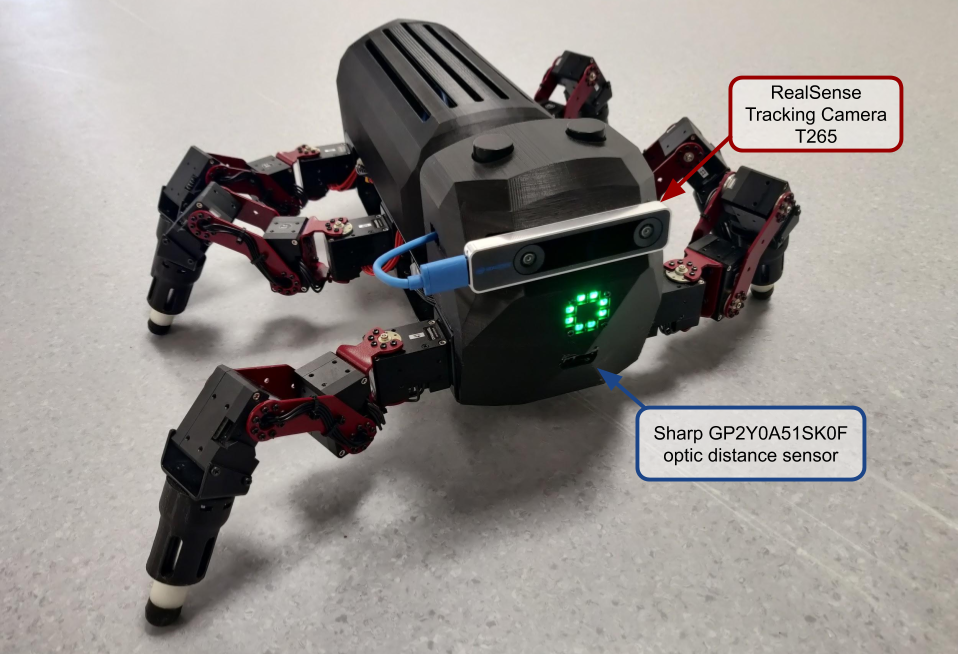}
  \caption{\small \textbf{The physical version of MORF}. MORF is equipped with a RealSense Tracking Camera T265 and a Sharp GP2Y0A51SK0F optic distance sensor. The learned closed-loop control modules require sensory information from these sensors to function.}
  \label{fig:physicalsetup}
\end{figure}

\begin{algorithm}[h]
\caption{$PI^{BB}$ Pseudocode}\label{alg:pibb}
\begin{algorithmic}[t]
\While{\textit{reward not converged}}
    \State \textit{// Execute $K$ roll-outs}
    \ForEach {$k \in K$}
        \State \textit{// Sample noise ($\epsilon_k$)}
        \State $\epsilon_k = \mathcal{N}(0,\sigma_{PI^{BB}}^2)$
        \State \textit{// Execute policy with noise and record final reward (R)}
        \State $R_{k} = \texttt{execCPGRBFN}(wp + \epsilon_k)$ 
    \EndFor
    \State \textit{// Calculate probability ($P_k$) for each roll-out}
    \ForEach {$k \in K$}
        \State $S_k = e^{\lambda\cdot \frac{R_k-\min_k(R_k)}{\max_k(R_k) - \min_k(R_k)}}$
    \EndFor
    \ForEach {$k \in K$}
        \State $P_k = \frac{S_k}{ \sum_{k=1}^{K}S_k}$
    \EndFor
\State \textit{// Perform reward-weighted averaging}
\State $\delta wp = \scriptstyle\sum_{k=1}^{K}\left(P_k\cdot \epsilon_k\right)$
\State \textit{// Update policy parameters}
\State $wp \gets wp + \delta wp$
\State \textit{// Decay exploration noise using $\gamma$}
\State $\sigma_{PI^{BB}}^2 = \gamma \cdot \sigma_{PI^{BB}}^2$
\EndWhile\label{euclidendwhile}
\end{algorithmic}
\end{algorithm}

\end{document}

%% file: tex/introduction.tex
Legged robots are mobile robots that have the potential to adapt to any environment on Earth accessible by their biological counterparts. They are flexible and not limited to paved or flat surfaces like wheeled robots~\cite{Leeeabc5986}. Legged robots are passively supported by the ground, enabling larger payloads with less effort when compared to flying robots \cite{Winkler2018}. Another advantage is their robustness to failures. Unlike wheeled and flying robots, legged robots often have a redundant number of legs and can, with the right control, continue to move even with several leg impairments \cite{Machado2006, CPGRBFN, Cully2015}. As a result of their many advantages, research on legged robots has grown significantly in the past decades~\cite{Silva2012}. Various kinds of legged robots with different morphologies have been developed and adopted for domains such as transportation, construction, exploration, inspection, and manipulation tasks \cite{Ansari,Centauro,spot,ANYbotics}. Legged robots are, however, still not able to fully explore and exploit their morphological potentials to achieve motion intelligence, like animals.

Today's approaches to developing legged robot control can be divided into two groups: model-based and model-free. Model-based approaches rely on analytical models describing the robot or system dynamics, which are tedious to construct and frequently inaccurate \cite{Hwangboeaau5872}. Furthermore, specialized control methods often need to be developed to tackle the complex problem of controlling legged robots, requiring a lengthy design process and manual parameter tuning \cite{MOMBAUR2017135}. Many model-based approaches apply a modular controller design whereby the controller is divided into smaller sub-modules that are decoupled and easier to design. For example, the popular control approaches presented in \cite{6943126,5175424,4115602} use template-dynamic-based sub-modules to approximate the robotic system as a point mass and calculate the next foothold or joint positions. The next sub-modules use these positions to compute trajectories to be followed. All sub-modules can be individually hand-tuned to adapt behavioral properties such as body elevation, step length, and step height. Despite these advantages, the approach is limited by the model accuracy and the fact that they are laborious. A controller must be manually developed, tuned, and tested, which often takes months \cite{Hwangboeaau5872}. Furthermore, this must be done for every new robot and task. 

On the other hand, model-free approaches can overcome many of the issues associated with model-based approaches by learning controllers directly from interacting with the environment using data-driven methods without the need for system or environment models. Many model-free approaches use reinforcement learning (RL) algorithms, often for policy optimization, where control parameters are tuned based on a reward function and extensive interactions with the environment. Model-free control thereby uses the fact that in many cases, especially for artificial legged locomotion, the model needed to predict and understand the physical dynamical system is a lot more complicated than the model needed to control the system. The controller is often implemented as a neural network where the network weights are used for control parameter optimization. Neural networks come in many sizes but are usually large (deep) and complex, with millions of weights to be learned. Large complex controllers, such as those presented in \cite{Leeeabc5986,Hwangboeaau5872,cluneencoding,schilling2020decentralized,Yangeabb2174,9224332}, display state-of-the-art locomotion control, but suffer from slow learning, with simulated learning times ranging from days to months. Simulated time is the time used by the robot inside the simulated environment when learning and therefore independent of the computer performance. The framed neural networks are also hard to comprehend, making it difficult to analyze and explain the learned control policy (non-explainable AI), thereby reducing user trust. Additionally, the complexity makes it hard to extend the framework with additional controllers in a modular way and the controllers often rely heavily on sensory information, which introduces a point of failure in cases of sensory fault. Sensory information is undoubtedly important for legged robot control and essential for adapting to unknown environments with difficult terrain. However, when sensory information is tightly coupled with the controller, it is difficult, and in some cases impossible, for the controller to continue operating in the case of sensory failures. On the other hand, the controller must not be too simple either. A simple controller, like those in \cite{multiopti,Cully2015}, does not facilitate complex locomotion control policies, thus limiting the performance of the robot.

\begin{figure}[!t]
\centering
\includegraphics[width=1\columnwidth]{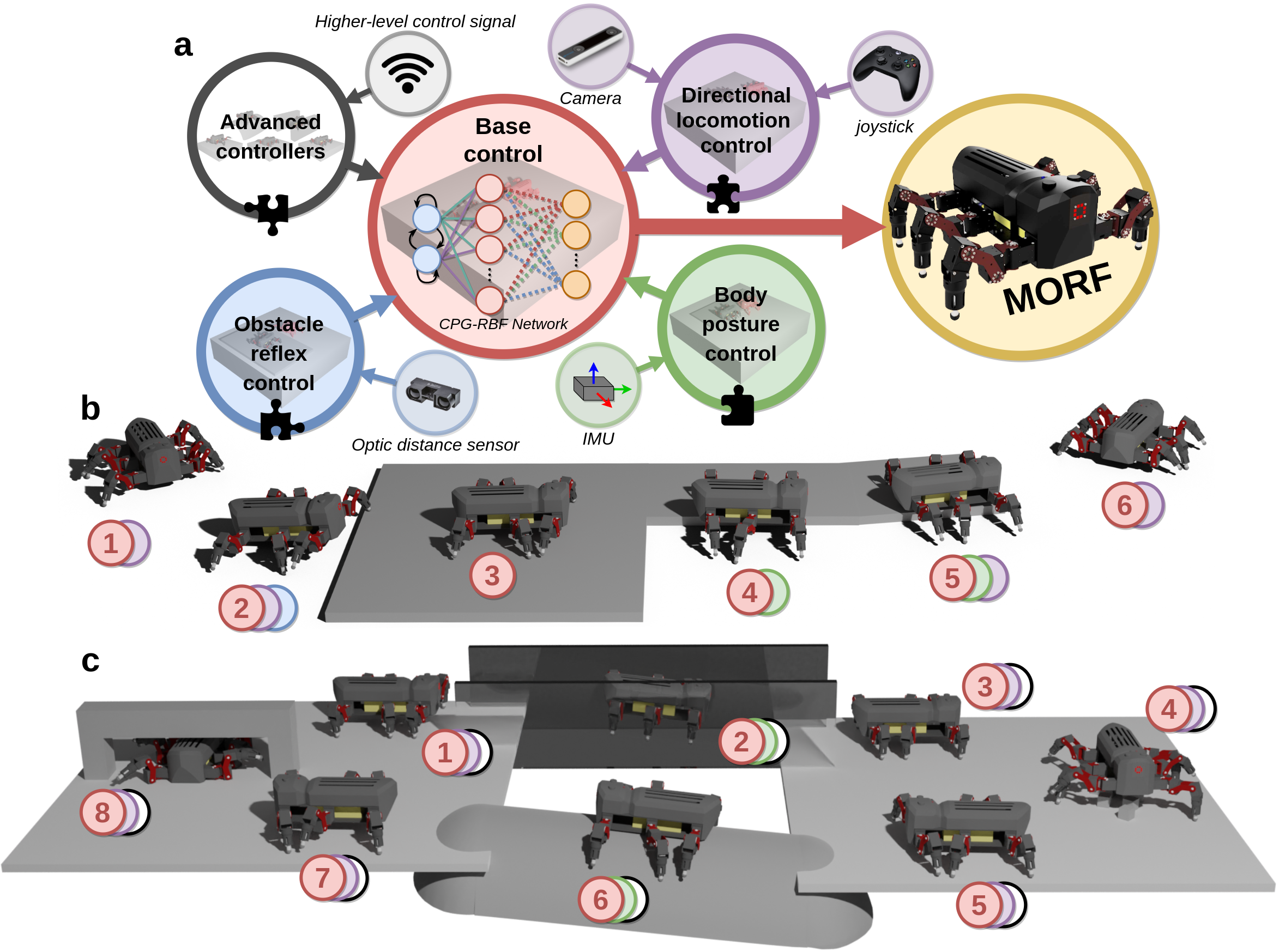}
\caption{\small \textbf{Overview of the versatile modular neural locomotion controller.} \textbf{a}, The controller comprises an open-loop base controller that constitutes the foundation for sub-behavior control modules encoding different robot behaviors/skills. \textbf{b-c}, The modules can be combined and used in parallel to enable a robot to traverse complex environments. In this work, we learn directional locomotion (purple circle), obstacle reflex (blue circle), body posture (green circle), and several advanced (dark gray circle) control modules. The base locomotion behavior (red circle) is encoded in the plastic synapses of the base controller shown with dashed lines in \textbf{a}, while the sub-behaviors are incrementally added to the controller in parallel through additional plastic synapses. In \textbf{b} and \textbf{c}, six and eight snapshots of a hexapod robot traversing a complex obstacle are shown. Each snapshot is indicated with circles, whose colors corresponds to the control modules being primarily used at that moment.}
\label{fig:simple_overview}
\end{figure}

To address the issues associated with current state-of-the-art approaches, we present a flexible modular neural controller with fast learning for motion intelligence\footnote{In this context motion intelligence refers to emergent, adaptive, and versatile locomotion behaviors that allow a legged robot to autonomously navigate in complex environments} of legged robots. The controller inherits all the advantages of model-free methods while addressing the problems of existing methods by being simple and easy to understand. Using this control approach, we demonstrate how a model-free approach allows fast learning of open- and closed-loop sub-behavior control modules, which encode different robot behaviors/skills (see Fig.\ \ref{fig:simple_overview}). Our controller is inspired by animal locomotion control principles, where locomotion is largely accomplished as an unconscious act. In principle, animal locomotion control comprises a genetically-encoded structure of the neural system such that animals can spend their first movement of life tuning the system instead of learning from scratch \cite{Kullander1889}. Biological studies has revealed that central pattern generators (CPGs) and premotor networks are encoded in these systems \cite{Kullander1889,animallocomotion,figurefeedback}. The CPG induces a natural gait with a strong prior on the agent's action space thereby significantly reducing the number of control parameters \cite{Azayev2020} while premotor networks reshape the CPG outputs. Inspired by this, the core of our controller combines a bio-inspired CPG with a premotor neural network into a so-called CPG-RBF network \cite{CPGRBFN} (see the base control in Fig.\ \ref{fig:simple_overview}a). Only a few plastic synapses (dashed lines in Fig.\ \ref{fig:simple_overview}a) needs to be learned to encode a base locomotion behavior. This locomotion behavior or base control module lays the foundation for the entire controller and does not rely on sensory information. As a result, it will continue to function even if all the sensors of the legged robot fail. 


Although CPGs can generate motor patterns without sensory feedback, it is still crucial when modulating the motor pattern to accommodate irregularities, such as obstacles and uneven terrain, orienting toward a goal, and orienting away from an obstacle. This ability is often referred to as "sensorimotor integration" and plays a major role in animal locomotion \cite{animallocomotion, figurefeedback}. Similar to animal locomotion, sensorimotor integration is also essential for legged robot control. Inspired by this, together with the fact that walking animals exhibit a modular organization of locomotion control elements \cite{Delcomyn1999}, we introduce primitive closed-loop sub-behavior control modules that can be added to the base controller (see Fig.\ \ref{fig:simple_overview}). The sub-behavior modules integrate sensory information and are, using a novel approach, implemented in parallel with the plastic synapses of the premotor network (for details, see Methods section). Sensory information is in this way able to modulate the output of the base controller based on the synaptic weights of their respective module. The primitive closed-loop modules can be quickly learned independently of each other using general objective functions and learning environments. The learned modules can afterward be added to the base controller and used in parallel without additional coordination mechanisms to achieve emergent locomotion control for complex environments requiring multiple sub-behaviors to navigate (see Fig.\ \ref{fig:simple_overview}b-c). Therefore, the ability to traverse complex environments is not considered a single behavior but as emergent behaviors derived from multiple sub-behaviors that are combined and activated when needed. To additionally show that the proposed control structure can also learn more advanced behaviors, the closed-loop modular structure is used to learn new advanced control modules that can be added in parallel to the existing base and closed-loop control modules. The advanced control modules can be activated by higher-level control signals from the user or higher-level control (see Fig.\ \ref{fig:simple_overview}a). In our approach, the robot has the ability to learn new and forget existing behaviors without destroying existing ones, which is a common problem of neural network learning \cite{Leeeabc5986,Hwangboeaau5872,cluneencoding,schilling2020decentralized,Yangeabb2174,9224332}. Forgetting behaviors is especially important in cases of sensory fault since those behaviors relying on broken sensors can be removed online. Finally, the separation of sensorimotor coordination into different control modules and the simplicity of the CPG-RBF network make the system very transparent and explainable (explainable AI \cite{Samek2019}).

In this work, we demonstrate how our control approach can be applied to a complex hexapod robot (called MORF \cite{MORF}) powered by 18 electric actuators. Firstly, an open-loop base controller is learned as a basis for robot walking. Three primitive closed-loop sub-behavior control modules for directional locomotion, obstacle reflex, and body posture stabilization are then incrementally added to the base controller. These controllers are learned separately in simple environments and later combined to obtain versatile emergent locomotion behaviors for traversing a complex environment both in simulation and the real world. Finally, we additionally learn five advanced control modules for locomotion at different heights (over and under obstacles), with narrow legs, on a pipe, and vertically between two walls. Together with the base and primitive modules, the five advanced control modules are used for traversing a highly complex environment in simulation. To this end, this study provides the following contributions beyond state of the art in robot locomotion control: 1) a simple modular neural controller, inspired by animal locomotion, which is embodied, scalable, transparent, and explainable; 2) fast and simple learning (within minutes) due to an encoded CPG-RBF network structure and few plastic synapses; 3) the ability to combine modules implementing different robot behaviors for locomotion in complex environments; 4) the ability to learn new and forget existing robot behaviors without compromising the rest of the controller.


%% file: tex/results.tex
\section*{Results}
Firstly, we demonstrate that our control structure can learn an open-loop base controller, enabling MORF to walk straight. We then present the results from learning three primitive closed-loop control modules added on top of the base controller (see Fig.\ \ref{fig:simple_overview}a). The first closed-loop control module implements obstacle reflex behavior, enabling MORF to negotiate obstacles in its path. The obstacle reflex controller uses a local optic distance sensor placed on the head of MORF. The second closed-loop control module implements body posture behavior through orientation sensory feedback from an inertial measurement unit (IMU) to minimize tilt movement. Finally, the third closed-loop control module implements directional locomotion behavior, enabling MORF to dynamically change its direction of motion.

Each controller is trained independently of each other on simple tasks and environments in simulation (see Supplementary Section S1 for simulation details). However, the controllers can later be combined and used in parallel for locomotion in complex environments. This is evidenced in a task wherein the base and three primitive controllers are used to traverse an environment with many obstacles and uneven terrain. This task is also validated on the physical MORF robot to demonstrate that the controller can be directly transferred to a physical system without any modification and operate in real-time. Through this task, we further show how our controller overcomes sensory fault by disabling and enabling the obstacle reflex control module online (Supplementary Section S3 and Video S9 shows the behavior when also disabling the other primitive modules one by one).

Finally, to demonstrate advanced emergent behaviors achieved by the proposed control approach, we present the results from learning five advanced control modules which are added on top of the base controller. The five additional modules can be activated and deactivated using higher-level control inputs and they enable MORF to perform advanced locomotion modes (see Methods section). In this work, the activation of modules is done manually, but it could easily be coupled with higher-level control. The first control module enables MORF to locomote with the body lifted high off the ground such that it can walk over obstacles in its way. The second module enables MORF to locomote with the body close to the ground such that it can squeeze under obstacles. The third module enables MORF to locomote with the legs close to the body such that it can fit in narrow spaces. The fourth module enables MORF to climb on pipes. The fifth module enables MORF to climb between walls vertically. In a final task, the base, three primitive, and five advanced controllers are merged to generate emergent adaptive behaviors to traverse a complex environment with many complex obstacles (see Fig.\ \ref{fig:simple_overview}c). The generalization and limitation of each primitive and advanced control module are investigated and discussed in Supplementary Section S2 and shown in Supplementary Video S8.

\subsection*{Results of learning the open-loop base controller}
\begin{figure}[t]
\centering
\includegraphics[width=1\textwidth]{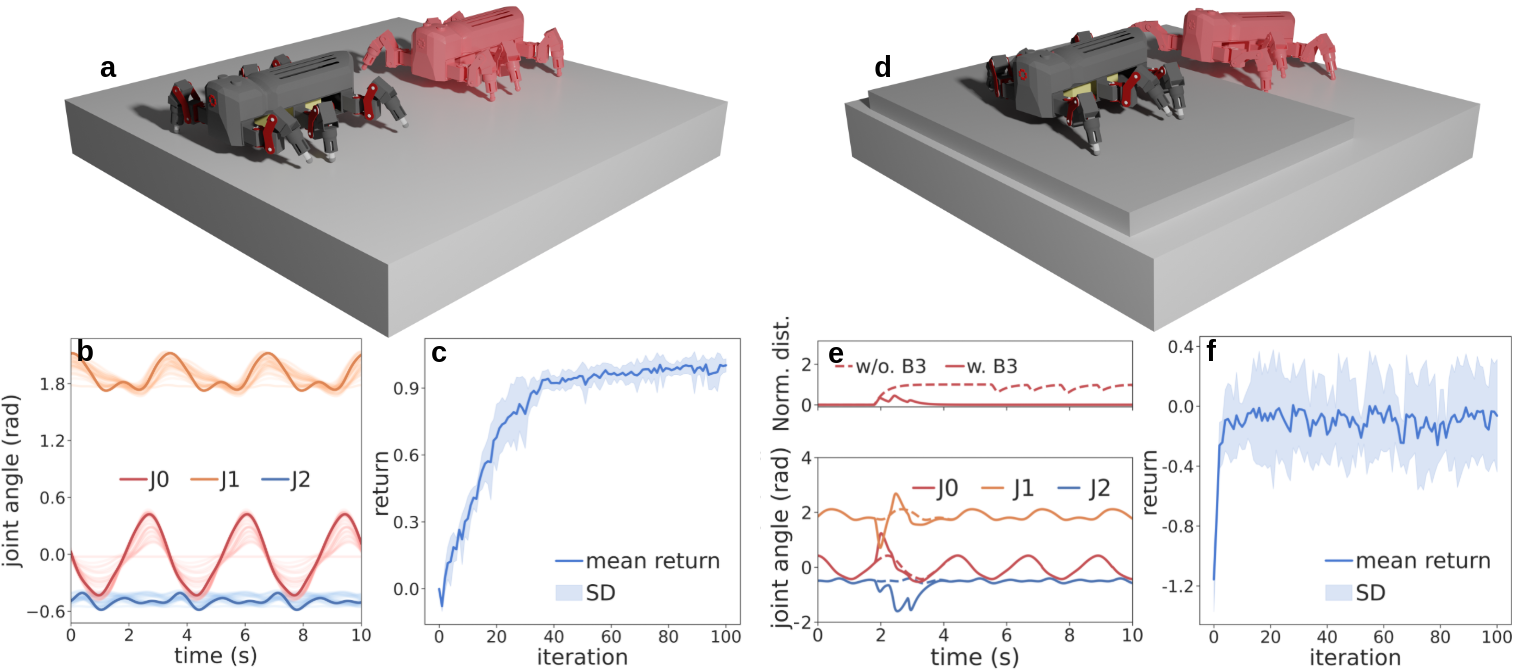}
  \caption{\small \textbf{Results of learning the base and obstacle reflex controllers}. \textbf{a-c}, results of learning the base controller. \textbf{a}, the simulated environment and learned base locomotion behavior -- the red MORF model shows an earlier time step. \textbf{b}, the learned leg joint trajectories (\textit{J0-2}) for a single leg. The solid lines show the converged trajectories, while the transparent lines show the intermediate joint trajectories starting from the first iteration. \textbf{c}, the mean and SD of the reward per iteration. \textbf{d-f}, results for learning the obstacle reflex controller. \textbf{d}, the simulated environment and learned obstacle reflex behavior -- the red MORF model shows an earlier time step. \textbf{e}, the top plot shows the normalized optic distance sensor values with (solid line) and without (dashed line) the obstacle reflex controller. The bottom plot shows the three leg joint trajectories (\textit{J0-2}) of a front leg with (solid line) and without (dashed line) the obstacle reflex controller. \textbf{f}, the mean and SD of the reward per iteration.}
  \label{fig:baseobs}
\end{figure}

The base controller implements an open-loop locomotion behavior and is the foundation for the eight additional control modules. The base locomotion behavior is encoded in the plastic synapses between the premotor network and the motor neurons (see Fig.\ \ref{fig:simple_overview}a). Each plastic synapse amplifies or suppresses specific parts of the periodic CPG output based on the activity of the corresponding RBF neuron and its weight. The outputs of the motor neurons are provided as position commands to the three leg joints \textit{J0}, \textit{J1}, and \textit{J2} in each of MORF's six legs (see Methods section). The base locomotion behavior is learned in the simulated environment shown in Fig.\ \ref{fig:baseobs}a. The learning process continues until the reward feedback converges, and a final set of weights for the behavior is learned. For the base locomotion behavior, the reward feedback is given by the distance walked and stability of the robot computed as the sum of variance in body yaw, pitch, and roll, as well as body height. Fig.\ \ref{fig:baseobs}b shows the resulting joint trajectories during learning, and Fig.\ \ref{fig:baseobs}c shows the mean and standard deviation (SD) of the reward for each iteration. The mean and SD are calculated over five learning sessions of 100 iterations each. Fig.\ \ref{fig:baseobs}b also shows that the reward feedback converges after 20 iterations or 16 minutes of simulated time. Supplementary Video S1 shows different iterations during the learning process.

\subsection*{Results of learning the primitive closed-loop control modules}
The three primitive closed-loop control modules are added on top of the base controller to modulate the already learned base locomotion trajectories. They are all encoded in plastic synapses that receive sensory feedback and runs in parallel with the synapses of the base controller (see Methods section). Each control module is learned in different simulated environments with the already learned base controller. The mean and SD of the reward are calculated over five learning sessions of 100 iterations each.

The first primitive control module implements an obstacle reflex behavior, using binary sensory feedback from an optic distance sensor placed on the head of MORF. The binary feedback is filtered using three low-pass single-pole infinite impulse response (IIR) filters in series. By placing the IIR filters in series, it is possible to add memory and consequently, retain the sensory feedback for a longer time. The control modules are trained in a simulated environment where MORF walks towards an obstacle in the form of a $0.04\,$m thick plate, as shown in Fig.\ \ref{fig:baseobs}d. The reward feedback is given as the distance walked as well as the stability of the robot. Fig.\ \ref{fig:baseobs}e shows the sensory feedback and joint trajectories of a single front leg with and without the learned reflex controller. Fig.\ \ref{fig:baseobs}f shows the mean reward, with the reward feedback converging within 10 iterations or 19 minutes of simulated time. The longer simulated time per iteration when compared to that of the base controller is caused by a longer roll-out length. Supplementary Video S2 shows different stages of learning the obstacle reflex control module.

\begin{figure}[t]
\centering
\includegraphics[width=1\textwidth]{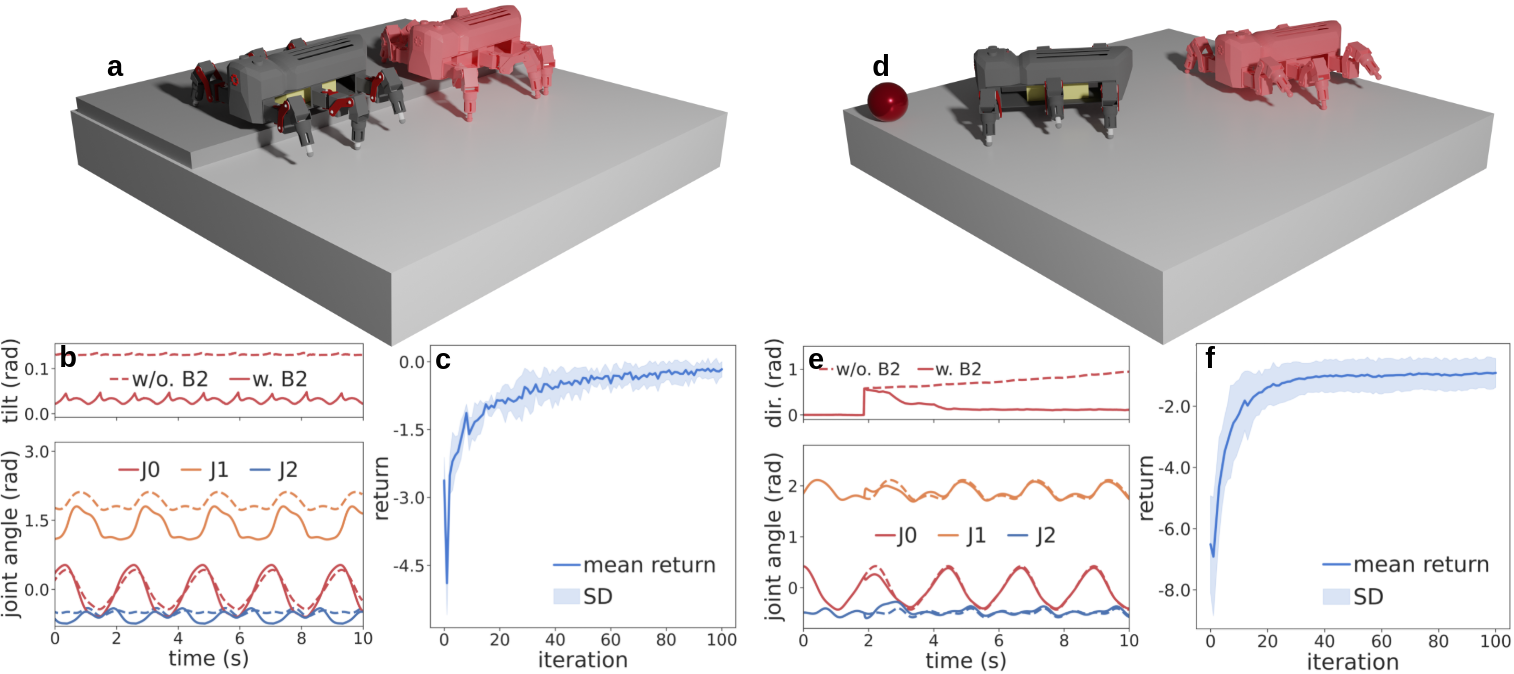}
  \caption{\small \textbf{Results of learning the body posture and directional locomotion controllers}. \textbf{a-c}, results for learning the body posture controller. \textbf{a}, the simulated environment and learned body posture behavior -- the red MORF model shows an earlier time step. \textbf{b}, the top plot shows the tilt sensor values with (solid line) and without (dashed line) the body posture controller. The bottom plot shows three leg joint trajectories (\textit{J0-2}) of a left leg with (solid line) and without (dashed line) the body posture controller. \textbf{c},~the mean and SD of the reward per iteration. \textbf{d-f}, results for learning the directional locomotion controller. \textbf{d}, the simulated environment and learned directional locomotion behavior -- the red MORF model shows an earlier time step. \textbf{e}, the top plot shows the heading direction error with (solid line) and without (dashed line) the directional locomotion controller. The bottom plot shows three leg joint trajectories (\textit{J0-2}) of a right leg with (solid line) and without (dashed line) the directional locomotion controller. \textbf{f}, the mean and SD of the reward per iteration.}
  \label{fig:tiltdir}
\end{figure}

The second primitive control module implements a body posture controller, using orientation sensory feedback from an IMU. The controller is trained in a simulated environment where MORF walks with its right legs on an obstacle in the form of a $0.04\,$m thick plate as shown in Fig.\ \ref{fig:tiltdir}a. The reward is given as the distance walked and the stability of the robot with an emphasis on avoiding tilting. Fig.\ \ref{fig:tiltdir}b shows the sensory feedback and joint trajectories of a right leg with and without the learned body posture controller. Fig.\ \ref{fig:tiltdir}c shows the mean reward for each iteration, with the reward feedback converging within 20 iterations or 16 minutes of simulated time. Supplementary Video S3 shows different stages of learning the body posture control module.

The third primitive control module implements a directional locomotion behavior, also using orientation sensory feedback from an IMU. The controller is trained in a simulated environment where a sphere will spawn after a few seconds, as shown in Fig.\ \ref{fig:tiltdir}d. The sensory feedback is then converted to the error between the heading direction of MORF and the direction of this sphere. The reward feedback is the distance walked, the robot's stability, and the error in heading direction. Fig.\ \ref{fig:tiltdir}e shows the heading direction error and joint trajectories of a left leg with and without the learned directional locomotion controller. Fig.\ \ref{fig:tiltdir}f shows the mean rewards for each iteration, with the reward feedback converging within 20 iterations or 24 minutes of simulated time. The longer simulated time per iteration is again due to an increased roll-out length. Supplementary Video S4 shows different stages of learning the directional locomotion control module.

\begin{figure}[ht]
\centering
\includegraphics[width=0.85\textwidth]{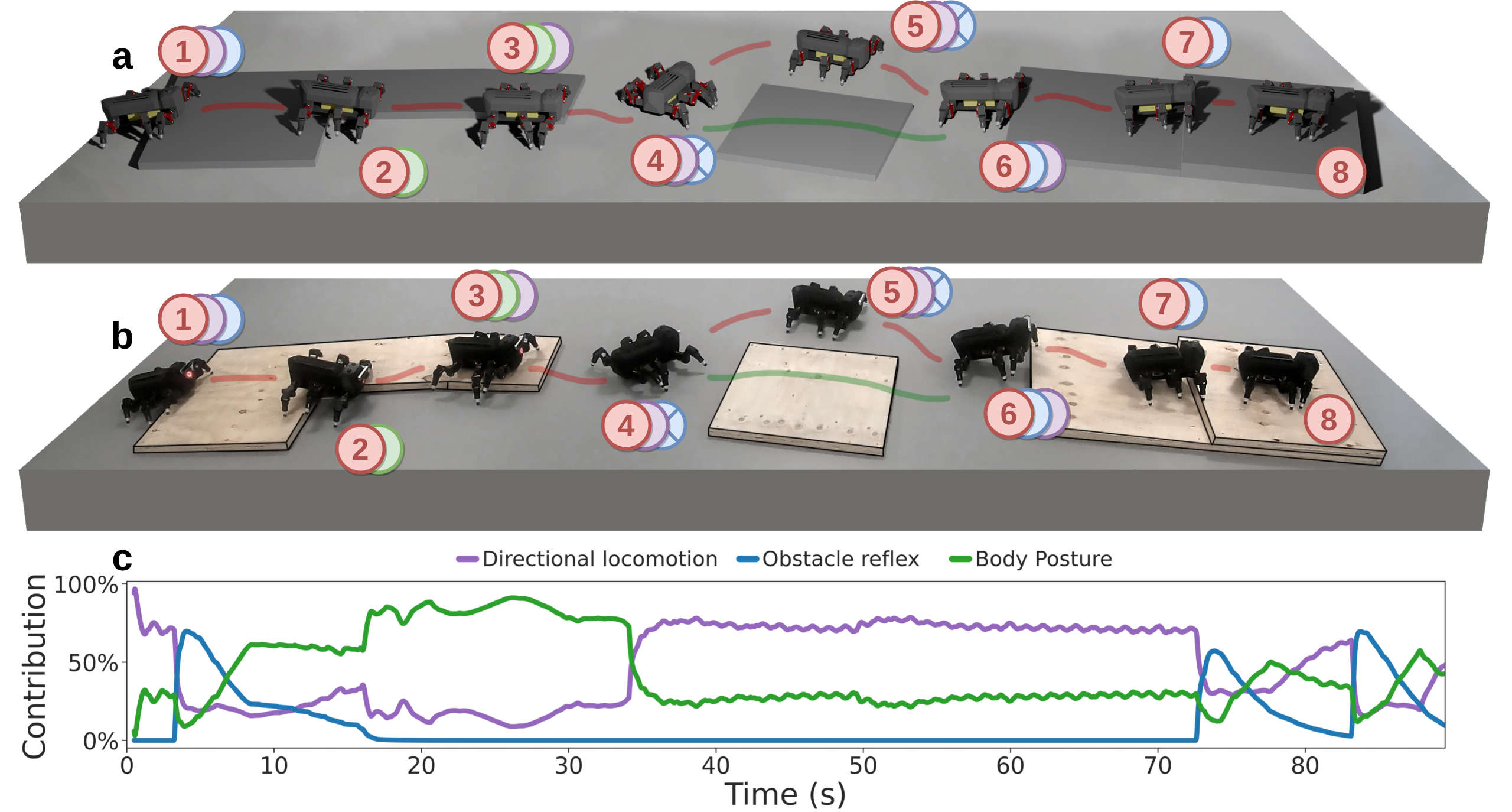}
  \caption{\small \textbf{Results for using the learned primitive control modules}. \textbf{a}, the simulated environment and eight snapshots of MORF when using the base controller (red circle), obstacle reflex controller (blue circle), body posture controller (green circle), and directional locomotion controller (purple circles). Each snapshot is indicated with circles, whose colors corresponds to the control modules being primarily used at that moment. Note that all control modules are active for the entire run with the exception of \raisebox{.5pt}{\textcircled{\raisebox{-.9pt} {4}}} and \raisebox{.5pt}{\textcircled{\raisebox{-.9pt} {5}}}, where the distance sensor placed on the head of MORF is temporally broken and the obstacle reflex controller disabled (blue circle with a cross). Consequently, MORF is steered around the obstacle (the red path) instead of traversing it (the green path). \textbf{b}, the real-world environment which is a copy of the simulated environment. The snapshot circles indicate the same as for the simulated case in \textbf{a}. The image is cropped to resemble the simulated environment, and the borders of the obstacles have been highlighted. For the original image, see Supplementary Video S6. \textbf{c}, the normalized contribution to the base controller from the three primitive control modules.}
  \label{fig:combi}
\end{figure}

\subsection*{Using all controllers and deploying on a physical robot}
One of the main advantages of modular neural controllers is that the learned controller modules can be used in parallel without any additional modification or network. Fig.\ \ref{fig:combi}a-b shows eight snapshots of how the base, obstacle reflex, body posture, and directional locomotion control modules can be used to traverse a complex environment both in simulation and on a physical robot in real-time. During walking, we simulate a temporary failure in the distance sensor placed on the head of MORF. Consequently, the obstacle reflex controller is manually swapped out, and MORF needs to rely on the other controllers to finish the task. In simulation (Fig.\ \ref{fig:combi}a), an alternative curving path (red path around the obstacle) is given such that MORF will take this path when the obstacle reflex controller is swapped out. In the real-world setup (Fig.\ \ref{fig:combi}b), the alternative curving path is provided by a human operator using a joystick and the camera on MORF. To later increase the autonomy of the robot, a fault detection algorithm, such as those presented in \cite{6864786,1415023,Christensen2008}, could be used to swap out the faulty control modules. The alternative path or desired walking direction could also be provided by high-level control algorithms, such as the path finding algorithms presented in \cite{PATLE2019582, pathfinding}. Fig.\ \ref{fig:combi}c shows how and when the three primitive control modules are activated and used in parallel. Here, all three primitive control modules are used in parallel on several occasions resulting in emergent behaviors. Supplementary Video S5 shows the simulated MORF traversing the course using all the controllers, while Supplementary Video S6 shows the performance of the controller on the physical MORF. The Supplementary Videos also show how MORF performs with fully functional sensors for the entire course (i.e., using the green path in Fig.\ \ref{fig:combi}a-b).

\subsection*{Results of learning advanced control modules}
\begin{figure}[t]
\centering
\includegraphics[width=0.88\textwidth]{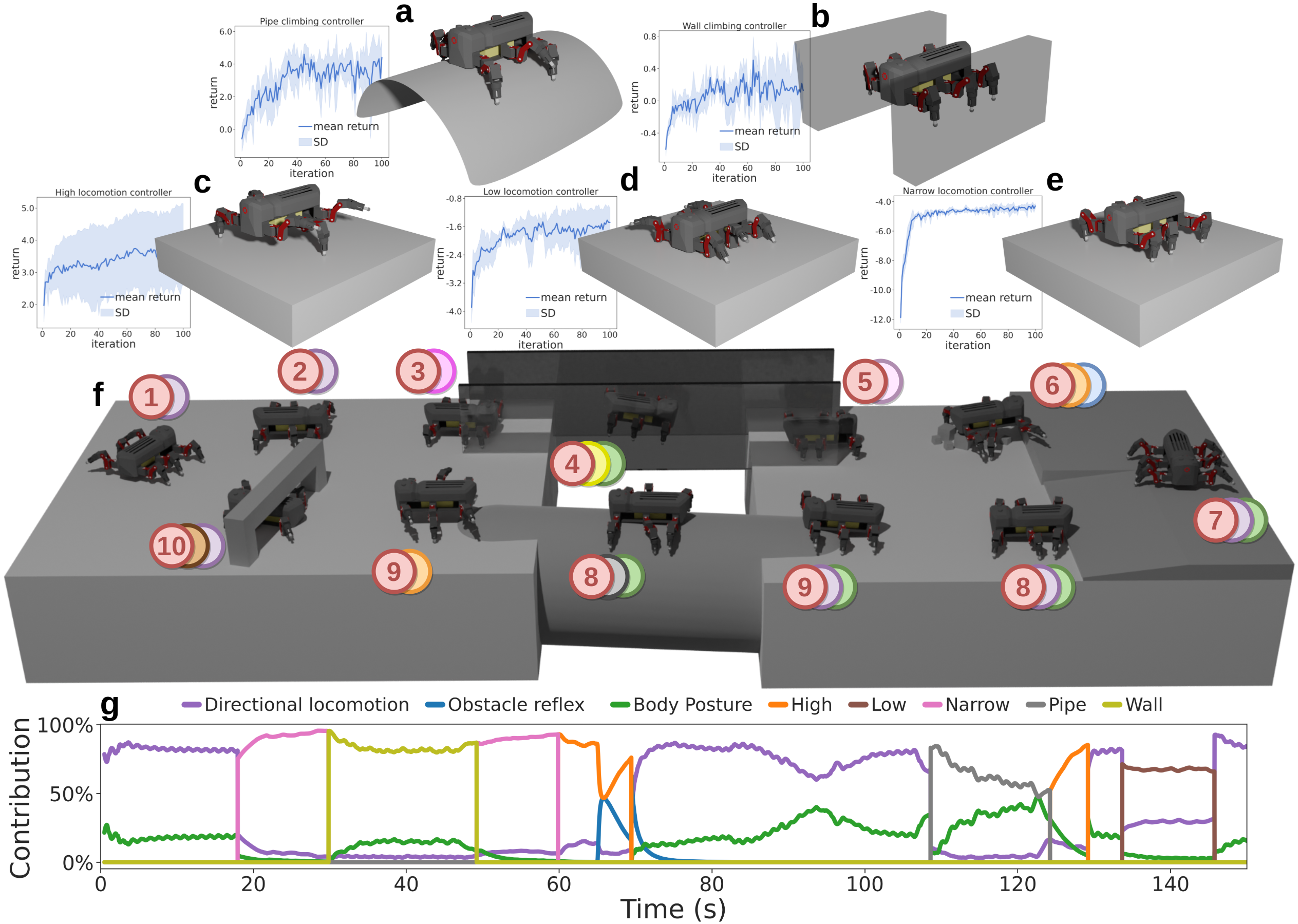}
  \caption{\small \textbf{The advanced behaviors and the results for using all eight control modules}. \textbf{a}, pipe climbing behavior. \textbf{b}, wall climbing behavior. \textbf{c}, high locomotion behavior. \textbf{d}, low locomotion behavior. \textbf{e}, narrow locomotion behavior. \textbf{a-e} also shows the mean and SD of the reward per iteration. \textbf{f}, the simulated environment and 10 snapshots of MORF when using the base controller (red circle), three primitive controllers (purple, blue, and green circles), pipe climbing controller (gray circle), wall climbing controller (yellow circle), high locomotion controller (orange circle), low locomotion controller (brown circle), and narrow locomotion controller (pink circle). Each snapshot is indicated with circles, whose colors corresponds to the control modules being primarily used at that moment. \textbf{g}, the normalized contribution to the base controller from the three primitive and five advanced control modules.}
  \label{fig:complexbehaviorsenv}
\end{figure}

Fig.\ \ref{fig:complexbehaviorsenv}a-e shows the five advanced behaviors included in this study together with the mean reward per iteration, where the reward feedback is converging within 28 minutes of simulated time on average. The five behaviors enable MORF to locomote on pipes, vertically between two walls, over and under obstacles, and in narrow spaces. Each control module is learned in different simulated environments with the already learned base controller. The mean and SD of the reward are calculated over five learning sessions of 100 iterations each. As discussed in Supplementary Section S2, the primitive closed-loop modules generalize to most of the behaviors produced by the advanced control modules. In many cases, the emergent behaviors even show better performance when compared to that of the advanced control module alone. By using both the advanced and primitive control modules MORF is able to adaptively traverse the complex environment shown in Fig.\ \ref{fig:complexbehaviorsenv}f and Supplementary Video S7. Fig.\ \ref{fig:complexbehaviorsenv}g shows when the eight modules are activated and used in parallel. Finally, Supplementary Video S10 shows the behavior when disabling each of the eight modules one by one.


%% file: tex/discussion.tex
\section*{Discussion}
The bio-inspired modular neural locomotion controller presented in this work comprises several independently learned control modules that encode different robot behaviors. We demonstrated that our novel approach can successfully learn an embodied open-loop base controller and, subsequently, learn three primitive closed-loop control modules as well as five advanced control modules that can adapt the base controller to complex environments. Our approach significantly reduces the complexity of learning controllers for legged locomotion by making them modular such that they can be learned sequentially. A modular setup has several advantages. Firstly, it is possible to use simple objectives and test environments where reward feedback convergence is more likely. Secondly, the controller can continue to grow with new modules or robot behaviors for specific environments as required. Control modules can likewise be removed, which is especially useful in cases of sensory faults. With only three layers, the neural control structure of our controller is also simple. As a result, only 60 learning parameters per control module (20 per joint) need to be learned, requiring 28 minutes of simulated learning time on average. These properties make the controller a promising method for future research on online learning on physical robots. The fast learning is due to the simplicity of the neural control architecture, making it easier to comprehend and explain, e.g., by analyzing how sensory feedback modulates the base locomotion trajectories. Finally, it was also demonstrated that the controller can be deployed on a physical system where the primitive control modules can be used in parallel without any additional modification. Note that the presented approach is not limited to the hexapod robot or environments used in this work. We consider the results presented in this paper as a step toward a general, scalable, analyzable, and explainable locomotion controller for complex legged locomotion control.

In \cite{Yangeabb2174}, a modular locomotion control architecture called multi-expert learning architecture (MELA) was recently presented. The MELA contains multiple expert neural networks, each with a unique motor behavior and gating neural network (GNN), fusing experts dynamically into a versatile and adaptive neural network. MELA uses a two-stage learning approach. In the first stage, expert network modules are trained on specific tasks, while in the second stage, all expert modules are co-trained with the GNN that learns how to blend the output of the experts for various tasks. Compared to a single complex deep neural network for locomotion control, as in \cite{Hwangboeaau5872}, the MELA architecture is more explainable and biologically plausible. However, each expert module, including the GNN, still use complex, hard to understand, deep neural network structures and, with the two-stage learning approach, require days of simulated time to optimize. Moreover, individual experts cannot be removed or added online since they are co-trained and needed by the GNN. Finally, all expert networks, including the GNN, tightly integrate sensory feedback. This makes the architecture vulnerable to sensory failure since the experts cannot be removed online.

Our modular setup is comparable to the subsumption architecture \cite{1087032}; a control architecture that couples sensory feedback which actions in an intimate, bottom-up fashion. It does so by dividing the complete behavior into sub-behaviors (or modules). These modules are then organized into layers, with each layer implementing a particular behavioral competence (e.g., explore environment, avoid obstacle, etc.). These layers are placed in a hierarchy where higher levels are able to combine or even inhibit the lower levels. In a similar way, some of the advanced control modules inhibit non-compliant primitive control modules (see Supplementary Section S2). An advantage of our approach is that while in the subsumption architecture, modules are typically designed manually, the modules are learned automatically in our approach. Finally, in both our controller and the subsumption architecture, the robot can still operate when some modules are missing, though with reduced capabilities.

As mentioned above, the three primitive closed-loop control modules presented in this work can be used in parallel without any additional modification. However, a limitation is that by continuing to add behavior-specific primitive modules to extend the controller's capabilities, different modules may start to interfere with one another. Like in the case of the advanced control modules, a solution is to use high-level control, like the subsumption architecture, such that modules can inhibit each other based on the overall objective. 

Even though the proposed control method allows for largely automated discovery of controllers, it still requires human expertise when designing training environments and reward functions. With a good understanding of the learning algorithm and task, similar to those presented in this work, the process of designing the rewards function and training environment as well as learning the control modules can take less than a day. To use the proposed controller for other simulated robots with different morphologies, modeling effort is required. The body and kinematics of the new robot must be modeled in simulation. Sensors and actuators likewise need to be set up, such that the relevant parameters are set accordingly. To account for modeling imprecision, noise can be applied to different aspects of the simulation to make the learned controller more robust. The general idea is to accept simulation imperfections while making the controller robust to variation \cite{radicalnoise}. Such robustness can be achieved by applying noise to the uncertain aspects of the simulation.

In summary, this work presents how simple neural open- and closed-loop control modules can be combined for complex locomotion control. Our approach utilizes some of the fundamental principles behind biological control, and can therefore serve as the basis for further studies on animal locomotion. Compared to state-of-the-art artificial locomotion methods, our method learns faster (reducing the learning time from days to minutes of simulated time), is simpler, and more flexible. 

%% file: tex/methods.tex
\section*{Methods}
\begin{figure}[!t]
\centering
\includegraphics[width=0.9\textwidth]{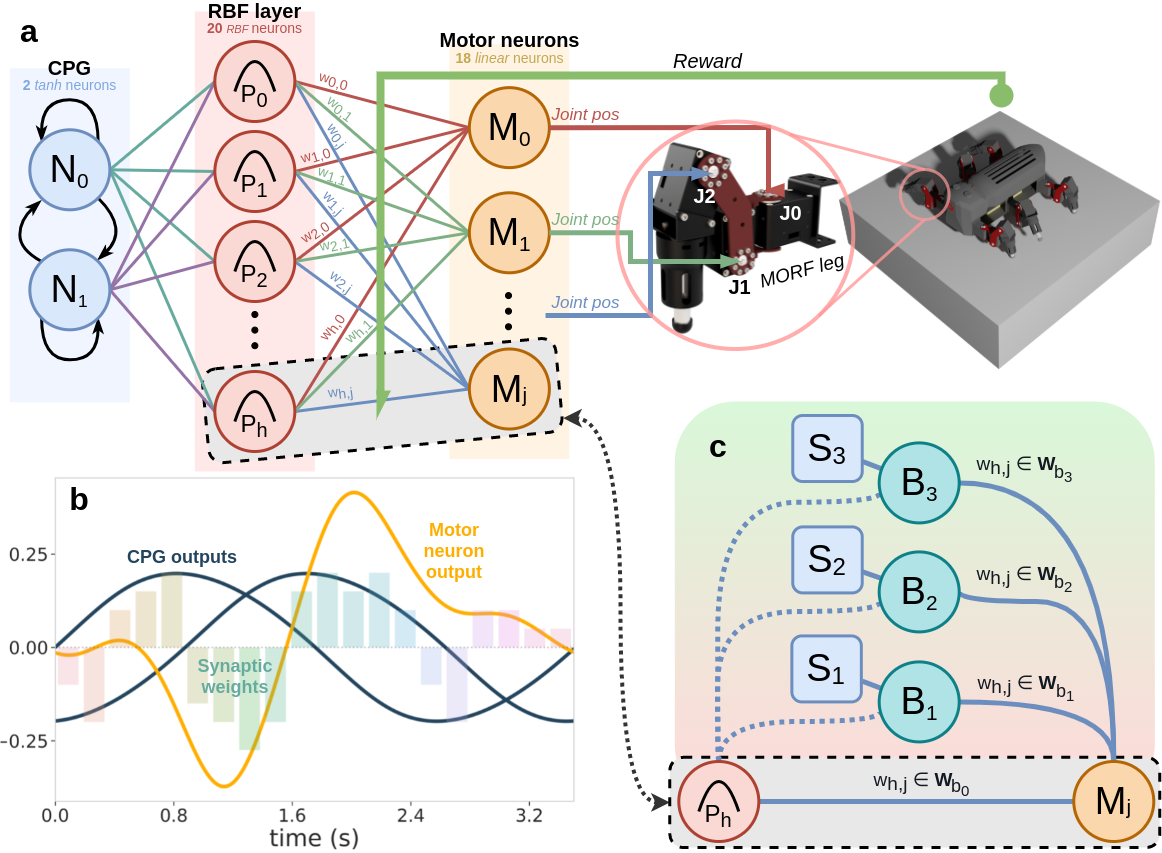}
  \caption{\small \textbf{Detailed overview of the versatile modular neural locomotion controller}. \textbf{a}, the CPG-RBF network consisting of a CPG ($N_{0-1}$), a premotor/RBF network ($P_{0-h}$), and motor neurons ($M_{0-j}$). The outputs of the motor neurons are target positions for the joints of the robot. The shape of the motor neuron output is encoded in the plastic synapses between the RBF layer and motor neurons. In other words, these plastic synapses encode the robot behaviors and are therefore optimized in the training loop. \textbf{b}, one period of the CPG outputs, randomly generated plastic synapses (bars in the plot), and motor neuron output. \textbf{c}, novel neural architecture enabling additional controllers or motor behaviors to be added in a modular way. The presented architecture consists of an open-loop base controller ($W_{b0}$) with three primitive closed-loop controllers on top for obstacle negotiation, body posture control, and directional locomotion control ($B_{1-3}$ or weight sets $W_{b_{1-3}}$), respectively. The three closed-loop controllers receive feedback from three sensors, where $S_1$ is the binary feedback from an optic distance sensor, $S_2$ is the tilt movement from an inertial measurement unit, and $S_3$ the error between the actual and desired heading direction. In simulation, the desired heading direction is provided by waypoints and on the real robot by a joystick.}
  \label{fig:overview}
\end{figure}

This section describes in detail the modular neural locomotion controller, learning process, and physical robot. An overview of the training loop is shown in Fig.\ \ref{fig:overview}. The CPG-RBF network that combines a CPG with a radial basis function (RBF) network was first introduced in \cite{CPGRBFN} as a single open-loop non-modular locomotion controller. The CPG outputs a rhythmic wave-shaped signal, and the RBF network, acting as a premotor network, reshapes this signal based on the robot morphology and desired behavior. In this work, we significantly extend the previous study and demonstrate that the controller can be expanded with primitive closed-loop and advanced control modules (see Supplementary Fig.\ \ref{fig:learning_approach}). These modules expand the capabilities of the controller, enabling legged robots to adapt in complex environments without compromising the simplicity or robustness of the control structure. 

\subsection*{Central pattern generator}
To generate the basic rhythmic signals for locomotion, we use a central pattern generator (CPG). A biological CPG is a cluster of nerve cells or interconnected neurons within the thoracic ganglia of invertebrates and spinal cord of vertebrates \cite{animallocomotion}. The CPGs play a key role in locomotion and other rhythmic movements \cite{Aoi2017,Nachstedt2017}. This is because they can generate motor patterns without requiring sensory feedback or any functional link to higher brain centers. In the CPG-RBF network, the abstract SO(2)-oscillator based artificial neural CPG model is used \cite{so2} (see the CPG in Fig.\ \ref{fig:overview}a). The SO(2)-oscillator is a neural network consisting of only two fully-connected standard additive discrete-time neurons ($N_{0-1}$), both using a sigmoid transfer function. The SO(2)-oscillator can produce rhythmic output signals with a phase shift of $\pi/2$ and display various dynamic behaviors (e.g., periodic patterns with varying frequencies, chaotic patterns, and hysteresis effects \cite{pasemann1998,pasemann2002,Steingrube2010}) by adjusting its synaptic weights through sensory feedback or manual control. These dynamical network behaviors can subsequently be exploited for complex locomotion modes (e.g., walking at different frequencies \cite{TNNLSDIL}, chaotic leg movement for self-untrapping of legs that are stuck \cite{Manoonpong2013}).

The two SO(2)-oscillator outputs are given by,
\begin{equation}
\begin{aligned}
    o_i(t+1)=\operatorname{tanh}\left (\sum_{j=0}^{N}w_{ij}(t)o_j(t)  \right ),
\end{aligned}
\label{eq:cpgoutput}
\end{equation}
\noindent where $o_i$ is the output of neuron $i$, $N$ is the number of neurons, and $w_{ij}$ is the weight of the synapses between neuron $i$ and $j$.
    
As proven by Pasemann et al. \cite{so2}, the network produces a quasi-periodic output when the weights are chosen as,
\begin{equation}
\begin{aligned}
    \begin{pmatrix}
    w_{00}(t) & w_{01}(t)\\ 
    w_{10}(t) & w_{11}(t) 
    \end{pmatrix} = \alpha \cdot
    \begin{pmatrix}
    \cos\:\varphi(t) & \sin\varphi(t)\\ 
    -\sin\:\varphi(t) & \cos\varphi(t)
    \end{pmatrix},
\end{aligned}
\label{eq:cpgweight}
\end{equation}
\noindent where $\varphi$ is the frequency-determining parameter, $\alpha$ determines the amplitude and the nonlinearity of the output oscillations. In this study, we use $\alpha = 1.01$ and $\varphi = 0.01\pi$ to obtain harmonic oscillation with a frequency of $\approx0.30$~Hz. The two SO(2)-oscillator outputs can be seen in Fig.\ \ref{fig:overview}b. Note that the frequency may be learned and optimized \cite{TNNLSDIL} together with the trajectory but is fixed for the purpose of this study.

\subsection*{Premotor network}
To reshape the otherwise fixed wave-shape output from the SO(2)-oscillator, a premotor network is used. As the premotor network, we use the RBF network; an artificial neural network with radial basis activation functions \cite{rbfn}. The RBF network is compact; consisting merely of a single hidden layer together with the input and output layer. For the CPG-RBF network, the CPG outputs form the input layer, while the motor neurons from the output layer (see Fig.\ \ref{fig:overview}). As mentioned above, the activation functions of the hidden neurons are radial basis functions, chosen in the case of the CPG-RBF network as two-dimensional Gaussian functions. RBF networks are commonly used in function approximation tasks and are, therefore, well suited to reshape the CPG outputs \cite{rbfn}. The reshaping is achieved by letting the activation of hidden neurons in the CPG-RBF network to encode the joint positions at a particular phase in the stepping cycle. In this way, the CPG-RBF network will be able to either amplify or suppress a particular part of the CPG signal and ultimately produce arbitrary shaped rhythmic joint target trajectories. The target trajectories are encoded in the plastic synapses connecting the hidden layer to the motor outputs (blue and red connections in Fig.\ \ref{fig:overview}). The activation functions of the hidden RBF neurons are given as,
\begin{equation}
\begin{aligned}
    P_h = e^{-\left(\frac{(o_{n_0}-\mu_{h,0})^2 + (o_{n_1}-\mu_{h,1})^2}{\sigma_{RBF}^2}\right)},
\end{aligned}
\label{eq:transferfunc}
\end{equation}
\noindent where, $\mu_{h,0}$ and $\mu_{h,1}$ are two means of RBF neuron $P_h$, $\sigma_{RBF}^2$ is the common variance for the two means, and $P_h$ is the response of the RBF neuron when receiving input $o_{n_0}$ and $o_{n_1}$ from the CPG. The means are manually set such that the activations of the hidden RBF neurons are uniformly distributed along one period of the CPG outputs. More specifically, the means are calculated as,
\begin{equation}
\begin{aligned}
    \mu_{h,x} = o_{n_x} \left (\frac{(h-1)\cdot T}{H-1} \right ),
\end{aligned}
\label{eq:mydist}
\end{equation}
\noindent where $x$ is the index of the CPG outputs, $T$ is the CPG signal period ($T\approx1/0.30\:$~Hz), and $H$ is the size of the hidden layer. By uniformly distributing the means along one CPG signal period it is possible to reshape parts of the CPG outputs without the means needing to be learned. For instance, when using $H=20$, it is possible to reshape the $j^{th}$ joint trajectory at the center of its rhythmic movement by altering synaptic weight $wp_{10,j}$ from the tenth hidden RBF neuron $P_{10}$ to motor neuron $M_j$.

The size of the hidden layer or amount of RBF neurons, $H$, directly correlates to the complexity of the target trajectory. A large hidden layer can facilitate complex trajectories that can approximate virtually any function, while a small hidden layer only can generate simple trajectories. However, a small hidden layer benefits from having few policy parameters and, consequently, a faster convergence rate. While the size of the hidden layer correlates with the trajectory's complexity, the common variance of the RBF neurons, $\sigma_{RBF}^2$, correlates with its smoothness. A high variance results in smooth trajectories, while a lower allows more high-frequent trajectories. A trade-off thus exists, and in this study, we experimentally set $\sigma_{RBF}^2=0.04$ and $H=20$, allowing the training of smooth and complex trajectories at reasonable convergence rates \cite{CPGRBFN}. Specifically, it was observed that a lower $\sigma_{RBF}^2$ resulted in jerky trajectories, while a higher one produced simplistic trajectories with low returns (i.e., the learned periodic trajectories had almost symmetrical ascending and descending slopes). For $H$, it was observed that a smaller hidden layer could not produce complex trajectories (e.g., periodic trajectories with asymmetrical ascending and descending slopes and different velocities as shown in Fig.\ \ref{fig:baseobs}b) as well as obtain high returns. In contrast, a larger hidden layer only reduced the convergence rate and did not improve the return.

Fig.\ \ref{fig:overview}c shows our novel way of expanding the CPG-RBF network with new behaviors and sensorimotor integration. The sensory feedback is integrated by introducing neurons ($B_{1-3}$) in parallel to the synapses connecting the RBF to the motor neurons. As explained above, the weights of these synapses encode the joint trajectories or base locomotion behavior. The parallel $B_{1-3}$ neurons are shunting inhibition neurons, set up such that their outputs are a multiplication of the RBF neuron ($P_h$) outputs and sensory input ($S_{1-3}$). The closed-loop controllers added on top of the base controller are encoded in the weights ($W_{b_{1-3}}$) of the synapses between the $B_{1-3}$ neurons and motor neurons. These weights therefore specify how much the sensory input changes or modulates the base joint trajectories at a particular phase in the stepping cycle. The motor neuron output can hereby be formulated as,
\begin{equation}
\begin{aligned}
    M_{j} =o_{P_{h}} \cdot W_{b_{0}} + \left ( \sum_{n=1}^{3}  o_{P_{h}} \cdot S_{n} \cdot W_{b_{n}} \right ),
\end{aligned}
\label{eq:output}
\end{equation}
\noindent where $o_{P_{h}}$ is the output from RBF neuron $P_h$.

\subsection*{Controller encoding}
Due to its flexibility, the CPG-RBF network can be implemented as either a centralized controller or a decentralized controller. When using the CPG-RBF network as a central controller, all legs will use the same joint trajectories. When the controller is decentralized, the individual legs or leg pairs will learn different trajectories and more complex control policies can be learned. However, decentralization comes at the expense of additional control policy parameters, and thus a lower convergence rate (for more details, see \cite{CPGRBFN}). In this study, we use the CPG-RBF network as a central controller not only because of its simplicity but also to demonstrate how fast the various controllers can be learned. In this way, we also lay the foundation for online optimization directly on physical robots.

\subsection*{Learning algorithm}
For learning the weights of the CPG-RBF network, we use $PI^{BB}$ \cite{stulphal}; a state-of-the-art learning mechanism. The $PI^{BB}$ is a probability-based black-box optimization (BBO) approach, following a direct policy search to update the policy parameters with respect to a reward function. The $PI^{BB}$ is a modified version of the RL-based method “policy improvement with path integrals” ($PI^{2}$) \cite{Theodorou2010}, using constant exploration and removing the need for temporal averaging. The changes introduced in $PI^{BB}$ result in a more simplistic algorithm, providing a faster convergence rate and higher accumulated reward (return) win comparison to $PI^{2}$. Moreover, $PI^{BB}$ is robust without the use of matrix inversions and can be applied to model-free learning problems with easy-to-construct reward function requirements \cite{PomaSnake}. The only open parameter of $PI^{BB}$ is the exploration noise \cite{stulphal}, and it has faster convergence when compared to other gradient-based RL approaches by one order of magnitude \cite{Theodorou2010}. The $PI^{BB}$ method was selected not only for the advantages mentioned above but also because $PI^{2}$, which is comparable to $PI^{BB}$, has been successfully applied in similar continuous, high-dimensional action spaces \cite{Theodorou2010,PomaSnake,stulphal}.

Supplementary Algorithm \ref{alg:pibb} shows the pseudocode of the $PI^{BB}$ algorithm. From this, it can be seen that $PI^{BB}$ executes $K$ roll-outs, with $K$ Gaussian exploration noise sets ($\epsilon_k$) added to the control policy parameter set. The outcomes of the $K$ roll-outs are $K$ rewards ($R_k$), representing how well the policy, with the added exploration noise sets, performed as determined by the reward function. Finally, the policy parameters are updated by calculating the probability for each roll-out and performing cost-weighted averaging.

For the CPG-RBF network, the control policy parameters are the plastic synapses between the hidden RBF neurons and motor neurons (i.e., $W_{b_{0-3}}$ in Fig.\ \ref{fig:overview}c). The exploration noise is task-specific, as specified in the following sections, and the number of roll-outs ($K$) is set to $8$ for all tasks. The exploration noise and number of roll-outs are both empirically selected to promote fast and stable learning. Besides being task-specific, the exploration noise is also linearly decayed during learning using a decay constant of $\gamma=0.995$ for all tasks. Decaying the exploration noise allows for large initial weight changes and enforces smaller changes or fine-tuning toward the end of the learning process.

\subsection*{Learning the base controller}
The base controller is open-looped and enables MORF to walk straight on flat ground with a stable body posture. It lays the foundation for additional control modules to be added on top (see Supplementary Fig.\ \ref{fig:learning_approach}). Since the CPG-RBF network is used as a central controller, a fixed leg phase relationship is employed whereby the contralateral legs operate in reciprocal fashion, resulting in a tripod gait. This behavior is also often seen in animals \cite{animallocomotion}. For learning the control parameters (i.e., weight set $W_{b_0}$ in Fig.\ \ref{fig:overview}) the following reward function is used,
\begin{equation}
\begin{aligned}
    R_k &= w_d\cdot d - ( w_\gamma\cdot \gamma + w_\xi \cdot \xi + w_\varsigma \cdot \varsigma) 
\end{aligned}
\label{eq:rewardfunction_base}
\end{equation}
\noindent where $w_d = 3$, $w_\gamma = 1$, $w_\xi = 3$, and $w_\varsigma = 0.75$. The reward function in equation (\ref{eq:rewardfunction_base}) consists of four sub-rewards: distance ($d$), instability ($\gamma$), body height error ($\xi$), and slippage ($\varsigma$). The distance sub-reward accounts for movement along the initial heading direction, and consequently, rewards fast straight locomotion. The instability sub-reward covers the stability of the robot during movement. It is computed as the sum of variance in body yaw (heading direction), pitch, and roll, as well as body height. A yaw of $0^{\circ}$ means that the robot is straight, whereas a pitch and roll equal to $0^{\circ}$ means that the robot is parallel with the floor. Instability, therefore, punishes movement that is not in the walking direction. The body height error sub-reward is a measure of the difference between mean body height during walking and desired walking height. The slippage sub-reward considers the extent to which each leg of the robot slips on the ground. The slippage return is calculated as the number of times a leg tip move (i.e., has a velocity greater than some threshold) while in ground contact. The slippage is normalized between 0 and 1 and the leg with the highest slippage is used as return. A slippage return of $1$ thus implies that one or more legs slip on the ground whenever in contact with it. Note that each sub-reward is multiplied by a weight ($w_d$, $w_\gamma$, $w_\xi$, and $w_\varsigma$) to make them similar in magnitude and range. The instability sub-reward is limited at $8$ to avoid negative returns becoming too large. In equation (\ref{eq:rewardfunction_base}), the distance sub-reward can be regarded as the dominating reward. Finally, the base controller is learned using an exploration noise ($\epsilon_k$) of $0.02$ and a roll-out execution time of six seconds.

\subsection*{Learning the obstacle reflex controller}
The obstacle reflex controller enables MORF to traverse obstacles in its way using sensory feedback from a binary optic distance sensor placed on the robot's head. The sensor uses a cutoff distance of $0.115$ m and is oriented $30$ degree forward from a downward facing position. The feedback is low-pass filtered by three IRR filters to retain the signal for a longer time and only applied to the two front legs. For learning the control parameters (i.e., weight set $W_{b_1}$ in Fig.\ \ref{fig:overview}), the following reward function is used,
\begin{equation}
\begin{aligned}
    R_k = w_d\cdot d - ( w_\gamma\cdot \gamma + w_\xi \cdot \xi + w_\varsigma \cdot \varsigma ).
\end{aligned}
\label{eq:rewardfunction_obstacle}
\end{equation}
where $w_d = 0.5$, $w_\gamma = 1$, $w_\xi = 0$, and $w_\varsigma = 0.5$. The reward function is similar to equation (\ref{eq:rewardfunction_base}) but with the exclusion of body height error ($\xi$) and a lower $w_d$ and $\varsigma$. The reason for excluding the body height error is that it will penalize the robot when crawling on top of an obstacle. Finally, the obstacle reflex controller is learned using an exploration noise ($\epsilon_k$) of $0.02$ and a roll-out execution time of $14$ seconds.

\subsection*{Learning the body posture controller}
The body posture controller enables MORF to keep a straight body by using sensory information from an onboard IMU. For simplicity, only sensory information from the x-axis orientation (i.e., tilting) is used. This sensory information can be seen directly as an error to be minimized whereby a tilt equal to zero means that the body is parallel with the ground. A negative tilt measurement is projected to the legs on the right side and a positive to the legs on the left side. Furthermore, the tilt error is low-pass filtered to remove unwanted sensory noise. For learning the control parameters (i.e., weight set $W_{b_2}$ in Fig.\ \ref{fig:overview}), the following reward function is used,
\begin{equation}
\begin{aligned}
    R_k = w_d\cdot d - ( w_\gamma\cdot \gamma + w_\xi \cdot \xi + w_\varsigma \cdot \varsigma + w_{\tau_\mu} \cdot \tau_\mu + w_{\tau_\sigma} \cdot \tau_\sigma).
\end{aligned}
\label{eq:rewardfunction_body}
\end{equation}
where $w_d = 2$, $w_\gamma = 1$, $w_\xi = 0$, $w_\varsigma = 0.5$, $w_{\tau_\mu} = 40$, and $w_{\tau_\sigma} = 10$. The reward function is similar to that expressed in equation (\ref{eq:rewardfunction_base}) but with the exclusion of body height error ($\xi$) and a lower $w_d$ and $\varsigma$. In addition, the tilt mean error ($\tau_\mu$) and variance ($\tau_\sigma$) are included as sub-rewards. Finally, the body posture controller is learned using an exploration noise ($\epsilon_k$) of $0.1$ and a roll-out execution time of six seconds.

\subsection*{Learning the directional locomotion controller}
The directional locomotion controller enables MORF to walk in a desired direction. The difference between the actual and desired walking direction provides a heading error. A negative heading error is projected to the legs on the right side and a positive to the legs on the left side. As in the previous cases, the heading error is low-pass filtered to remove sensory noise. In both the simulation and in the physical robot, the actual walking direction is the robot's z-axis orientation (i.e., yaw) provided by the onboard IMU. In the simulation, the desired walking direction is provided by waypoints that can be arbitrarily placed in the scene. On the physical robot, the desired walking direction is provided by a joystick that, when combined with the onboard camera, can be used by a human operator to steer the robot even if outside the field of vision. For learning the control parameters (i.e., weight set $W_{b_3}$ in Fig.\ \ref{fig:overview}), the following reward function is used,
\begin{equation}
\begin{aligned}
    R_k = w_d\cdot d - ( w_\gamma\cdot \gamma + w_\xi \cdot \xi + w_\varsigma \cdot \varsigma + w_\delta \cdot \delta).
\end{aligned}
\label{eq:rewardfunction_direction}
\end{equation}
where $w_d = 0.1$, $w_\gamma = 1$, $w_\xi = 3$, $w_\varsigma = 1$ and $w_\delta = 6$. The reward function is similar to that expressed in equation (\ref{eq:rewardfunction_base}) with exception of a lower $w_d$ and higher $w_\varsigma$. In addition, the heading error ($\delta$) is included as a sub-reward. Finally, the directional locomotion controller is learned using an exploration noise ($\epsilon_k$) of $0.02$ and a roll-out execution time of $10$ seconds.

\subsection*{Learning the advanced controllers}
The five advanced controllers are integrated into the controller using the same novel architecture as the primitive closed-loop control modules (see Fig.\ \ref{fig:overview}). Instead of sensory feedback ($S_{1-3]}$), the advanced control modules received higher-level control signals that activates or disables the respective module. In this work, a simple binary decision input from an operator is used. The binary input signal is projected to all joint in all controller and may alternatively be provided by either higher-level control as well. The advanced control modules are learned in a similar way to the three primitive behaviors. All five controller are learned using an exploration noise ($\epsilon_k$) of $0.06$ but uses different rewards functions and roll-out execution time as explained in the following paragraphs.

\paragraph{The high locomotion controller} is learned using a roll-out execution time of $6$ seconds and the following reward function,
\begin{equation}
\begin{aligned}
    R_k = w_d\cdot d - ( w_\gamma\cdot \gamma + w_\varsigma \cdot \varsigma + w_{\lambda_y} \cdot \lambda_y + w_{\lambda_z} \cdot \lambda_z).
\end{aligned}
\label{eq:rewardfunction_high}
\end{equation}
where $w_d = 3$, $w_\gamma = 10$, $w_\xi = 0$, $w_\varsigma = 1$, $w_{\lambda_y} = -2$, and $w_{\lambda_z} = -15$. The reward function is similar to that expressed in equation (\ref{eq:rewardfunction_base}) with exception of a higher $w_\varsigma$ and removal of the body height error ($\xi$). In addition, the minimum body width ($\lambda_y$) and height ($\lambda_z$) are included as sub-rewards.

\paragraph{The low locomotion controller} is learned using a roll-out execution time of $6$ seconds and the following reward function,
\begin{equation}
\begin{aligned}
    R_k = w_d\cdot d - ( w_\gamma\cdot \gamma + w_\varsigma \cdot \varsigma + w_{\lambda_z} \cdot \lambda_z).
\end{aligned}
\label{eq:rewardfunction_low}
\end{equation}
where $w_d = 3$, $w_\gamma = 10$, $w_\xi = 0$, $w_\varsigma = 1$, and $w_{\lambda_z} = 60$. The reward function is similar to that expressed in equation (\ref{eq:rewardfunction_high}) with exception of a higher $w_{\lambda_z}$ and removal of the minimum body width ($\lambda_y$) measurement.

\paragraph{The narrow locomotion controller} is learned using a roll-out execution time of $6$ seconds and the following reward function,
\begin{equation}
\begin{aligned}
    R_k = w_d\cdot d - ( w_\gamma\cdot \gamma + w_\varsigma \cdot \varsigma + w_{\lambda_y} \cdot \lambda_y).
\end{aligned}
\label{eq:rewardfunction_narrow}
\end{equation}
where $w_d = 3$, $w_\gamma = 10$, $w_\xi = 0$, $w_\varsigma = 1$, and $w_{\lambda_y} = 60$. The reward function is similar to that expressed in equation (\ref{eq:rewardfunction_high}) with exception of a higher $w_{\lambda_y}$ and removal of the minimum body height ($\lambda_z$) measurement.

\paragraph{The pipe climb controller} is learned using a roll-out execution time of $14$ seconds and the following reward function,
\begin{equation}
\begin{aligned}
    R_k = w_d\cdot d - ( w_\gamma\cdot \gamma + w_\varsigma \cdot \varsigma + w_{\lambda_y} \cdot \lambda_y).
\end{aligned}
\label{eq:rewardfunction_pipe}
\end{equation}
where $w_d = 5$, $w_\gamma = 1$, $w_\xi = 0$, and $w_\varsigma = 0.75$. The reward function is similar to that expressed in equation (\ref{eq:rewardfunction_base}) with exception of a higher $w_{d}$, a lower $w_\varsigma$ and removal of the body height error ($\xi$).

\paragraph{The wall locomotion controller} is learned using a roll-out execution time of $12$ seconds and the following reward function,
\begin{equation}
\begin{aligned}
    R_k = w_d\cdot d - ( w_\gamma\cdot \gamma + w_\varsigma \cdot \varsigma + w_{d_z} \cdot d_z).
\end{aligned}
\label{eq:rewardfunction_wall}
\end{equation}
where $w_d = 0.6$, $w_\gamma = 1$, $w_\xi = 0$, $w_\varsigma = 0.75$, and $w_{d_z} = -6$. The reward function is similar to that expressed in equation (\ref{eq:rewardfunction_base}) with exception of a lower $w_d$ and $w_\varsigma$ as well as the and removal of the body height error ($\xi$). In addition, the distance traveled upwards ($d_z$) is included as a sub-reward.

\subsection*{Robot platform}
We used the MORF robot \cite{MORF}, shown in Fig.\ \ref{fig:simple_overview}a, to demonstrate the real-world applicability of our method. MORF is $0.42$ m long hexapod robot weighing about $4.2$ kg. Each leg is about $0.25$ m long when fully stretched and has three actuated degrees of freedom (DOFs), resulting in a total of 18 DOFs. MORF is equipped with Dynamixel XM430 coreless electric motors set to position mode using the built-in PID controllers. 

In both the simulation and on the physical robot, the controller was executed at $60$ Hz. A joystick was used with the physical robot as input to the directional locomotion controller modules together with visual feedback from the RealSense Tracking Camera T265. The RealSense Tracking Camera T265 also provide the tilting information for the body posture controller. Finally, the Sharp GP2Y0A51SK0F was used as an optic distance sensor for the obstacle reflex controller. The physical MORF with the sensors attached can be seen in the Supplementary Fig.\ \ref{fig:physicalsetup}.